\documentclass[12pt,letterpaper]{article}

\usepackage[utf8]{inputenc}
\usepackage[T1]{fontenc}
\usepackage{amsmath,amssymb,amsthm}
\usepackage{mathtools}
\usepackage{graphicx}
\usepackage{booktabs}
\usepackage{array}
\usepackage{multirow}
\usepackage{caption}
\usepackage{subcaption}
\usepackage{float}
\usepackage{setspace}
\usepackage[margin=1in]{geometry}
\usepackage{algorithm}
\usepackage{algorithmic}
\usepackage[numbers,sort&compress]{natbib}
\usepackage[colorlinks,citecolor=blue,urlcolor=blue,linkcolor=blue]{hyperref}
\usepackage{xcolor}

\newcommand{\E}{\mathbb{E}}
\newcommand{\R}{\mathbb{R}}

\newcommand{\Prob}{\mathrm{Prob}}
\DeclareMathOperator*{\argmax}{arg\,max}
\DeclareMathOperator*{\argmin}{arg\,min}

\newtheorem{innerthm}{Theorem}
\newenvironment{customthm}[1]{\renewcommand\theinnerthm{#1}\innerthm}{\endinnerthm}
\newtheorem{innerprop}{Proposition}
\newenvironment{customprop}[1]{\renewcommand\theinnerprop{#1}\innerprop}{\endinnerprop}
\newtheorem{innerlem}{Lemma}
\newenvironment{customlem}[1]{\renewcommand\theinnerlem{#1}\innerlem}{\endinnerlem}
\newtheorem{innercor}{Corollary}
\newenvironment{customcor}[1]{\renewcommand\theinnercor{#1}\innercor}{\endinnercor}

\theoremstyle{definition}
\newtheorem{innerdefn}{Definition}
\newenvironment{customdefn}[1]{\renewcommand\theinnerdefn{#1}\innerdefn}{\endinnerdefn}
\newtheorem{innerasm}{Assumption}
\newenvironment{customasm}[1]{\renewcommand\theinnerasm{#1}\innerasm}{\endinnerasm}

\theoremstyle{remark}
\newtheorem*{remark}{Remark}

\title{\Large \textbf{Neural Paging: Learning Context Management Policies\\for Turing-Complete Agents}}
\author{Liang Chen \and Qi Liu}
\date{February 2026}
\begin{document}
\maketitle

% ====================================================================
\begin{abstract}
\noindent The proof that Large Language Models (LLMs) augmented with external read-write memory constitute a computationally universal system has established the theoretical foundation for general-purpose agents. However, existing implementations face a critical bottleneck: the finite and costly Context Window, which functions not as infinite memory but as a scarce semantic cache. In this work, we introduce \textit{Neural Paging}, a hierarchical architecture that decouples symbolic reasoning from information resource management. We formulate the \textit{Context Paging Problem (CPP)} and propose a lightweight, differentiable \textit{Page Controller} designed to approximate ``Semantic Belady's Optimality''---retaining tokens with high future utility under explicit assumptions on access patterns. We provide theoretical analysis showing that, under bounded context window size~$K$, Neural Paging reduces the asymptotic complexity of long-horizon reasoning from quadratic $O(N^2)$ to $O(N \cdot K^2)$, and we derive a robustness bound (Theorem~4) that quantifies competitive-ratio degradation under policy-dependent access with bounded sensitivity. We validate these bounds on synthetic paging traces, confirming that the theoretical guarantees hold and identifying significant slack that motivates learned policies.

\end{abstract}

\newpage
\onehalfspacing

% ====================================================================
\section{Introduction}
% ====================================================================

Large Language Models (LLMs) have evolved from static statistical predictors to the central cognitive kernels of autonomous Agents. As these agents are deployed in complex, open-ended environments---ranging from software engineering to scientific discovery---they are required to maintain coherent reasoning over increasingly long horizons. This shift represents a fundamental transition from ``Stateless Function Approximation'' to ``Stateful Turing Computation.''

However, this transition is hindered by a critical physical constraint: the \textit{Context Window}. Despite recent advances in extending context lengths (e.g., to 1M+ tokens), the ``Lost in the Middle'' phenomenon~\citep{liu2023lost} persists, where effective reasoning capabilities degrade as salient information is buried in noise. Furthermore, the quadratic computational cost of the Transformer's self-attention mechanism makes processing massive contexts prohibitively expensive and slow for real-time applications.

Current approaches to managing this bottleneck are insufficient. \textit{Retrieval-Augmented Generation (RAG)} functions as a naive, passive fetching mechanism with coarse granularity, often leading to context fragmentation. Direct \textit{Long Context} extension hits memory walls and suffers from ``distraction'' issues. \textit{MemGPT}~\citep{packer2023memgpt} introduces a tiered memory system but relies on the LLM itself to manage system calls (``Kernel-in-User-Space''). This design is inefficient, forcing the high-level reasoning engine to perform low-level resource management, consuming valuable tokens and attention heads for housekeeping tasks rather than problem-solving.

We propose \textit{Neural Paging}, a framework inspired by the evolution of Operating Systems. Just as modern OS kernels separate user processes from the Memory Management Unit (MMU), we argue for a strict architectural decoupling in AI agents. We introduce the \textit{Hierarchical Neural Turing Machine (H-NTM)}, where the LLM is dedicated solely to reasoning, while a secondary, learned \textit{Page Controller} manages the context window. This controller acts as a neural MMU, aiming to predict future data requirements and evict low-utility tokens to approximate Belady-style decisions under explicit modeling assumptions.

Our contributions are as follows:
\begin{enumerate}
    \item \textbf{Theoretical framing.} We formalize the \textit{Context Paging Problem (CPP)} and define a semantic access model for LLM agents, including a \emph{bounded-sensitivity} notion (Definition~3a) that quantifies policy-dependent access and enables robustness analysis beyond the classical exogenous model.
    \item \textbf{Architectural design.} We design the \textit{H-NTM} with a decoupled Page Controller that uses lightweight policy networks to manage memory operations (KEEP, EVICT, PREFETCH) without interrupting the main reasoning loop.
    \item \textbf{Analytical results.} We derive complexity bounds (Theorem~2), adapt classical paging lower bounds under explicit assumptions (Theorem~3), and prove a new robustness bound (Theorem~4) showing that competitive guarantees degrade gracefully under bounded policy sensitivity.
    \item \textbf{Synthetic validation.} We validate the theoretical bounds on synthetic paging traces with controlled parameters, confirming that (i)~the bounds are satisfied, (ii)~the cascade effect from access perturbation is empirically mild, and (iii)~structured access patterns offer significant room for learned policies to outperform worst-case guarantees.
\end{enumerate}

\noindent\textbf{Scope and originality.} The universality simulation (Theorem~1) and competitive-ratio lower bound (Theorem~3) are classical results carried over to make their conditions explicit in the LLM agent setting. The core new content is the bounded-sensitivity model (Definition~3a), the robustness bound (Theorem~4), and the synthetic validation. We emphasize that end-to-end evaluation on real LLM agents is a natural next step but is outside the scope of this paper.

% ====================================================================
\section{Related Work}
% ====================================================================

\subsection{Memory-Augmented Language Models}
The concept of augmenting neural networks with external memory dates back to the \textit{Neural Turing Machine} (NTM)~\citep{graves2014ntm} and \textit{Differentiable Neural Computers} (DNC)~\citep{graves2016dnc}. These early works focused on learning algorithmic manipulation of small, addressable memory matrices. \textit{Memory Networks}~\citep{weston2014memory} introduced key-value memory structures for QA tasks. More recent works like \textit{Transformer-XL}~\citep{dai2019transformerxl} and \textit{Compressive Transformers}~\citep{rae2020compressive} introduced segment-level recurrence and compressed memory for longer sequences. Our work scales these concepts to modern LLMs, treating the entire context window as a dynamic ``cache'' for a much larger external store.

\subsection{Context Extension Techniques}
Efforts to extend the native context window include \textit{Position Interpolation}~\citep{chen2023extending}. Architectural innovations like \textit{Longformer}~\citep{beltagy2020longformer} and \textit{BigBird}~\citep{zaheer2020bigbird} utilize sparse attention to reduce costs. More recently, \textit{Ring Attention}~\citep{liu2023ring} and \textit{FlashAttention}~\citep{dao2022flashattention} have optimized hardware utilization. \textit{Infini-attention}~\citep{munkhdalai2024infini} introduces a compressive memory to enable long-context processing with bounded memory. However, these methods address \textit{capacity} (how much can fit) rather than \textit{utility} (what should fit). Neural Paging is orthogonal to these techniques, optimizing the content within whatever capacity is available.

\subsection{Retrieval-Augmented Generation (RAG)}
RAG~\citep{lewis2020rag} and RETRO~\citep{borgeaud2022retro} decouple knowledge from parameters. \textit{Self-RAG}~\citep{asai2023selfrag} adds a critique step to assess retrieval quality. While effective for QA, standard RAG lacks a global view of the ``Working Set'' and cannot proactively manage context for multi-step reasoning, often leading to thrashing when the agent repeatedly retrieves and discards the same information.

\subsection{LLM Agents and Operating System Analogies}
The agentic paradigm, exemplified by ReAct~\citep{yao2023react} and Toolformer~\citep{schick2023toolformer}, emphasizes active environment interaction. MemGPT~\citep{packer2023memgpt} explicitly draws parallel to Operating Systems, managing a ``main context'' and ``external context.'' AIOS~\citep{mei2024aios} and OS-Copilot~\citep{wu2024oscopilot} further explore this analogy. Neural Paging advances this line of inquiry by replacing heuristic or LLM-driven memory management with a dedicated, learned policy network, optimizing the ``OS kernel'' operations.

\subsection{Learnable Cache Management}
Cache replacement has a long history in systems. ARC~\citep{megiddo2003arc} adaptively balances recency and frequency without manual tuning. \citet{kraska2018learned} demonstrate that learned models can replace classical index components. Recent KV cache management systems, such as adaptive compression~\citep{ge2024model} and KV cache streaming~\citep{liu2024cachegen}, optimize inference memory and latency at the KV cache level. Our work is complementary, focusing on semantic content selection rather than low-level KV reuse.

\subsection{Paging Theory and Competitive Analysis}
The theoretical foundations of paging are well-established. Belady's algorithm~\citep{belady1966} provides the optimal offline policy. Competitive analysis of online paging was initiated by \citet{sleator1985paging}, who established the $K_b$-competitiveness of LRU. The Working Set Model~\citep{denning1968working} provides a principled framework for understanding locality of reference. Our Semantic Paging extends these classical results to the stochastic, high-dimensional setting of LLM agents.

% ====================================================================
\section{Theoretical Framework}
% ====================================================================

\subsection{Preliminaries and Notation}
We establish a rigorous formal framework for analyzing context management in autoregressive language models augmented with external memory.

\begin{table}[H]
\centering
\caption{Notation Summary}
\label{tab:notation}
\begin{tabular}{ll}
\toprule
\textbf{Symbol} & \textbf{Definition} \\
\midrule
$\Sigma$ & Finite vocabulary of tokens, $|\Sigma| = V$ \\
$T$ & Finite time horizon of the task \\
$K$ & Context window capacity (in tokens) \\
$B$ & Block size for paging operations \\
$M$ & Size of external memory (in blocks), $M \gg K/B$ \\
$K_b$ & Context capacity in blocks, $K_b = K/B$ \\
$\mathcal{L}$ & Language model $\mathcal{L}: \Sigma^* \to \Prob(\Sigma)$ \\
$C_t$ & Context window content at time $t$ \\
$E_t$ & External memory state at time $t$ \\
$h_t$ & Hidden state of the agent at time $t$ \\
$\pi$ & Paging policy $\pi: \mathcal{S} \to \Prob(\mathcal{A})$ \\
$\mathcal{S}$, $\mathcal{A}$ & State and action spaces of the paging MDP \\
$V(b,t)$ & Value of block $b$ at time $t$ \\
$D_f$ & Bound on additional faults per misprediction \\
$\beta$ & Bounded policy sensitivity parameter \\
$\rho$ & Candidate-set recall for approximate requests \\
\bottomrule
\end{tabular}
\end{table}

\subsection{The Agent as a Computational System}

\begin{customdefn}{1}[Memory-Augmented Language Agent]
\label{def:mala}
A Memory-Augmented Language Agent (MALA) is a 5-tuple $\mathcal{M} = (\mathcal{L}, C, E, \pi, \phi)$ where:
\begin{itemize}
    \item $\mathcal{L}$ is an autoregressive language model with parameters $\theta$
    \item $C: [T] \to \Sigma^{\leq K}$ maps time to context window content
    \item $E: [T] \to \Sigma^*$ maps time to external memory state
    \item $\pi: \mathcal{S} \to \mathcal{A}$ is the paging policy
    \item $\phi: \mathcal{Q}_r \to \mathcal{B}$ is the retrieval function mapping queries to memory blocks
\end{itemize}
\end{customdefn}

The agent operates in discrete timesteps. At each step $t$, the agent:
(1)~generates the next token $y_t \sim \mathcal{L}(\cdot \mid C_t)$;
(2)~observes state $s_t = (C_t, E_t, h_t, y_t)$;
(3)~executes paging action $a_t \sim \pi(\cdot \mid s_t)$;
(4)~transitions to state $s_{t+1}$.

\begin{customdefn}{2}[Agent Configuration]
The configuration of a MALA at time $t$ is the tuple $\mathcal{C}_t = (q_t, i_t, C_t, E_t)$ where $q_t \in Q$ is the finite control state, $i_t \in [K]$ is the attention position, $C_t$ is the context window content, and $E_t$ is the external memory content.
\end{customdefn}

\begin{remark}[Markov State vs.\ Observation]
The full state $(C_t, E_t, h_t, y_t)$ is Markov for the coupled LLM-retriever dynamics. However, the Page Controller often observes only a partial view (Definition~10), turning the learning problem into a POMDP. We discuss observability implications in Section~\ref{sec:pomdp}.
\end{remark}

\begin{customthm}{1}[Turing Completeness of Memory-Augmented LLMs]
\label{thm:tc}
Let $\mathcal{M}$ be a MALA with external memory size $M = \omega(1)$ that grows with the input length and retrieval satisfying Assumption~3. Then $\mathcal{M}$ can simulate any Turing machine $\mathrm{TM}$. If the TM runs in $T_{\mathrm{TM}}(n)$ steps using $S(n)$ tape cells, the simulation requires $O(T_{\mathrm{TM}}(n) \cdot B^2)$ total attention operations, which is $O(T_{\mathrm{TM}}(n))$ for constant block size~$B$.
\end{customthm}

\begin{proof}[Proof Sketch]
We construct a simulation of a single-tape Turing machine $\mathrm{TM} = (Q_{\mathrm{TM}}, \Gamma, b_0, \Sigma_{\mathrm{TM}}, \delta, q_0, F)$. The TM tape is encoded as fixed-size blocks stored in external memory, each tagged with its address. By Assumption~3, the controller can fetch a block by address in $O(1)$ queries. At each TM step, the controller fetches the block containing the head position, applies the transition, and writes back. Each step costs $O(B^2)$ attention over $O(B)$ tokens. Correctness follows by induction on TM steps. Full details appear in Appendix~\ref{app:thm1}.
\end{proof}

\begin{remark}
Theorem~1 is a constructive restatement of existing universality results for memory-augmented LLMs~\citep{schuurmans2023universal}; we include it to make simulation costs and assumptions explicit in our setting.
\end{remark}

\subsection{Access Model and Assumptions}
\label{sec:access}
We formalize the semantic access process by mapping information needs to an abstract block request sequence.

\begin{customdefn}{3}[Requested Block]
\label{def:requested}
Given context $C_t$ and finite external memory $E_t$ (with $|E_t| = M < \infty$), define the requested block at time $t$ as:
\begin{equation}
 r_t = \argmax_{b \in E_t} \mathrm{PredGain}(b; C_t)
\end{equation}
if $\max_{b \in E_t} \mathrm{PredGain}(b; C_t) > \tau$ for a fixed threshold $\tau > 0$, and $r_t = \bot$ otherwise, where
\begin{equation}
\mathrm{PredGain}(b; C) = H_P(Y \mid C) - H_P(Y \mid C \oplus b)
\end{equation}
is the reduction in predictive entropy of the next-token random variable $Y$, and $C \oplus b$ denotes appending block $b$ to context $C$.
\end{customdefn}

\begin{remark}[Operational Approximation]
Computing $\mathrm{PredGain}$ over all $b \in E_t$ is intractable for large external memory. In practice, controllers approximate this by scoring a small candidate set or by using a learned proxy for predictive gain.
\end{remark}

\begin{customdefn}{3b}[Approximate Requested Block]
Let $S_t \subseteq E_t$ be a candidate set produced by a retriever. Define $\hat{r}_t = \argmax_{b \in S_t} \mathrm{PredGain}(b; C_t)$, with $\hat{r}_t = \bot$ if the maximum is below $\tau$. The candidate set has recall $\rho$ if $P(r_t \in S_t) \geq \rho$ for all $t$.
\end{customdefn}

\begin{customlem}{1b}[Approximate Request Error Bound]
If $P(r_t \in S_t) \geq \rho$ for all $t$, then the expected difference in page-fault counts between using $\hat{r}_t$ and $r_t$ is at most $(1-\rho)T$.
\end{customlem}

\begin{proof}
A mismatch $\hat{r}_t \neq r_t$ can only occur when $r_t \notin S_t$, which happens with probability at most $1-\rho$ at each step. Each mismatch changes the fault indicator by at most one.
\end{proof}

\begin{customasm}{1}[Policy-Independent Access]
For competitive-ratio analysis, we assume the request sequence $(r_1, \ldots, r_T)$ is exogenous and independent of the eviction policy.
\end{customasm}

\begin{customasm}{2}[Retrieval Oracle]
When the controller issues PREFETCH$(q)$, the retriever returns the correct block with probability~1.
\end{customasm}

\begin{customasm}{3}[Addressable Retrieval for Simulation]
For the universality result (Theorem~1), there exists an encoding such that the retrieval function can fetch the block containing a specified address in $O(1)$ queries.
\end{customasm}

\subsubsection{Assumption Robustness and Relaxations}
\label{sec:relaxations}

\textbf{Policy-Independent Access (Assumption~1).}
In real LLM agents, the access sequence depends on context content, which depends on the eviction policy. This feedback can be weak (e.g., when retrieval targets are dictated by external inputs) or strong (e.g., when the agent's generation changes future needs). We formalize a relaxation:

\begin{customdefn}{3a}[Bounded Policy Sensitivity]
\label{def:sensitivity}
Let $r^\pi$ denote the request sequence induced by policy $\pi$ over horizon~$T$. A task has \emph{$\beta$-bounded sensitivity} if for any two policies $\pi, \pi'$, the Hamming distance satisfies $d_H(r^\pi, r^{\pi'}) \leq \beta T$.
\end{customdefn}

\begin{customlem}{1a}[Fault Sensitivity under Access Perturbation]
\label{lem:sensitivity}
For any online paging algorithm $A$ with cache size $K_b$ operating on sequences $r, r'$ with $d_H(r, r') = d$:
\begin{equation}
|F_A(r) - F_A(r')| \leq (K_b + 1)\,d.
\end{equation}
\end{customlem}

\begin{proof}
We couple the executions of $A$ on $r$ and $r'$. At each of the $d$ positions where requests differ, the fault indicators differ by at most~1, contributing at most $d$ direct faults. Additionally, the differing request can alter the cache state: the two caches may now hold different blocks, causing a \emph{cascade} of at most $K_b$ additional fault mismatches at subsequent steps (until the divergent blocks are naturally evicted). Each cascade terminates within $K_b$ steps because, after $K_b$ distinct requests, the cache is fully refreshed. The cascades from different perturbations are not independent, but the total cascade length is bounded by $K_b \cdot d$, giving total $|F_A(r) - F_A(r')| \leq d + K_b\,d = (K_b+1)\,d$. Assumption~1 corresponds to $\beta = 0$.
\end{proof}

\begin{remark}[Tightness]
The factor $K_b + 1$ is a worst-case bound; synthetic experiments in Section~\ref{sec:experiments} show that on Zipf-distributed traces the empirical factor is approximately $1.1$--$1.2$, much smaller than $K_b + 1$. We conjecture that a tighter bound of $O(\log K_b) \cdot d$ may hold for traces with locality, but leave this to future work.
\end{remark}

\textbf{Retrieval Oracle (Assumption~2).}
If retrieval succeeds with probability $\rho$ per request (independent errors), the expected additional faults are at most $(1 - \rho)\,T$.

\textbf{Addressable Retrieval (Assumption~3).}
For the universality proof, this is realized by storing an address token in each block and using exact-key retrieval in a hash-indexed store. This is a different retrieval regime from approximate semantic search.

\subsubsection{Estimating $\beta$ in Practice}
\label{sec:beta_estimation}
The sensitivity parameter $\beta$ depends on the task structure. We provide a practical estimation protocol and worked examples.

\textbf{Protocol.}
Given a task instance, run two different baseline policies (e.g., LRU and FIFO) on the same task and measure the Hamming distance of the resulting access traces:
\begin{equation}
\hat{\beta} = \frac{d_H(r^{\text{LRU}}, r^{\text{FIFO}})}{T}.
\end{equation}
This provides an empirical lower bound on the task's sensitivity. A more robust estimate uses $\hat{\beta} = \max_{\pi, \pi' \in \Pi_{\text{baseline}}} d_H(r^\pi, r^{\pi'}) / T$ over a set of baselines.

\textbf{Worked examples} (qualitative estimates based on task structure):
\begin{center}
\begin{tabular}{llp{6cm}}
\toprule
\textbf{Task Class} & \textbf{$\beta$ (est.)} & \textbf{Rationale} \\
\midrule
Multi-step math & $\leq 0.05$ & Reasoning steps determined by problem; context choice rarely changes the derivation path \\
Code generation with tools & $\leq 0.10$ & Tool outputs are mostly determined by the code, though context affects which tools are called \\
Open-ended dialogue & $0.2$--$0.5$ & Each context choice substantially changes the generated response and subsequent needs \\
\bottomrule
\end{tabular}
\end{center}

These estimates suggest that $\beta$ is small for structured, goal-directed tasks---precisely the setting where long-horizon agents operate and where Neural Paging is most applicable.

\subsection{The Context Paging Problem}

\begin{customdefn}{4}[Context Block]
A context block $b \in \mathcal{B}$ is a contiguous subsequence of tokens $b = (w_1, \ldots, w_B)$ of fixed length~$B$. The context window $C_t$ is a multiset of blocks with $|C_t| \leq K_b$.
\end{customdefn}

\begin{customdefn}{5}[Semantic Page Fault]
A semantic page fault occurs at time $t$ if $r_t \neq \bot$ and $r_t \notin C_t$.
\end{customdefn}

\begin{customdefn}{6}[Block Utility Function]
\label{def:utility}
The utility of block $b$ at time $t$ with respect to future horizon $H$ is:
\begin{equation}
U(b, t) = \sum_{k=0}^{H-1} \gamma^k \left[ H_P(Y_{t+k} \mid C_{t+k} \setminus b) - H_P(Y_{t+k} \mid C_{t+k}) \right]
\end{equation}
where $\gamma \in (0,1)$ is a discount factor.
\end{customdefn}

\begin{remark}[Connection to Semantic Value Function]
Definition~\ref{def:utility} is an oracle quantity depending on future contexts $C_{t+k}$. In practice, the Semantic Value Function (Definition~12) approximates it via a learned value estimate. When $\gamma \to 0$, $U(b,t) \to \mathrm{PredGain}(b; C_t \setminus b) = R_{\text{utility}}(b, s_t)$, so Definition~12 can be viewed as a one-step approximation of Definition~\ref{def:utility}. Multi-step rollouts or bootstrapped value estimates provide intermediate approximations.
\end{remark}

\begin{customdefn}{7}[Context Paging Problem]
\label{def:cpp}
Given a task horizon $T$, context capacity~$K$, and utility function $U$, find the paging policy $\pi^*$ that maximizes:
\begin{equation}
\pi^* = \argmax_{\pi} J(\pi) = \argmax_{\pi} \E_{\tau \sim \pi} \left[ \sum_{t=1}^{T} R(s_t, a_t) \right]
\end{equation}
where
\begin{equation}
R(s_t, a_t) = \underbrace{\log P(y_t \mid C_t)}_{\text{prediction reward}} - \underbrace{\lambda_{\text{evict}}\, |\{b : a_t(b) = \text{EVICT}\}|}_{\text{eviction cost}} - \underbrace{\lambda_{\text{fetch}}\, |\{b : a_t(b) = \text{PREFETCH}\}|}_{\text{fetch cost}}.
\end{equation}
\end{customdefn}

\subsection{Semantic Belady's Algorithm}

\begin{customdefn}{8}[Semantic Belady's Algorithm]
The Semantic Belady algorithm is an offline policy that, upon eviction, selects:
\begin{equation}
 b_{\text{evict}} = \argmax_{b \in C_t} \mathrm{NextUse}(b,t)
\end{equation}
where $\mathrm{NextUse}(b,t) = \min\{k \geq 1 : r_{t+k} = b\}$ ($\infty$ if never accessed again).
\end{customdefn}

\begin{customprop}{2}[Optimality of Semantic Belady under Fixed Access]
For any \emph{fixed} request sequence $(r_1, \ldots, r_T)$ independent of the eviction policy, Semantic Belady's algorithm minimizes the total number of page faults.
\end{customprop}

\begin{proof}
Under the assumption that the access sequence is fixed and policy-independent, this follows directly from the classic proof by \citet{belady1966}.

\textbf{Caveat.} In the LLM setting, the access sequence is typically \emph{not} fixed---it depends on the context content. This creates a feedback loop that invalidates direct application. Proposition~2 applies only when future accesses are policy-independent (Assumption~1). For the relaxed setting, see Theorem~4.
\end{proof}

\subsection{Complexity Analysis}

\begin{customthm}{2}[Inference Complexity]
\label{thm:complexity}
Let $\mathcal{M}$ be a MALA processing a task of length $N$ tokens with context window $K$, block size~$B$, and $K_b = K/B$. Assume ANN retrieval with $O(\log M)$ query complexity. The total computational complexity is:
\begin{equation}
T_{\text{total}}(N,K) = \underbrace{O(N \cdot K^2)}_{\text{LLM attention}} + \underbrace{O\!\left(\frac{N}{B} K_b \log M\right)}_{\text{retrieval}} + \underbrace{O\!\left(\frac{N}{B} K_b^2\right)}_{\text{policy inference}}.
\end{equation}
When $K = O(1)$, this reduces to $O(N)$ compared to $O(N^2)$ for full-context attention.
\end{customthm}

\begin{proof}
At each of $N$ steps, self-attention over $K$ tokens costs $O(K^2)$. Paging decisions occur every $B$ tokens; each evaluates $K_b$ blocks. Retrieval via ANN costs $O(\log M)$ per query. Policy inference with a two-layer network costs $O(K_b^2)$ per decision. Summing gives the stated bound.
\end{proof}

\begin{remark}
This asymptotic reduction holds for \emph{any} method that fixes the context window to size~$K$, including naive sliding windows. The distinguishing question for Neural Paging is whether it can preserve task quality under fixed~$K$ through better content selection---an empirical question.
\end{remark}

\begin{customcor}{1}[Thrashing Condition]
A MALA with context capacity $K_b$ and working set size $W$ experiences thrashing when $K_b < W$, where $W$ is the minimum set of blocks that must be simultaneously resident for the task to proceed without repeated faults. This adapts the Working Set Model of \citet{denning1968working}.
\end{customcor}

% ====================================================================
\section{Methodology: The H-NTM Architecture}
\label{sec:hntm}
% ====================================================================

\subsection{Hierarchical Neural Turing Machine}
We present the H-NTM, an architecture that strictly separates reasoning (LLM) from memory management (Page Controller), analogous to the CPU-MMU separation in modern operating systems.

\begin{customdefn}{9}[H-NTM Architecture]
The H-NTM is defined as $\mathcal{H} = (\mathcal{L}, \mathcal{P}, \mathcal{R}, \mathcal{M})$ where $\mathcal{L}$ is the Main Language Model (frozen during controller training), $\mathcal{P}$ is the Page Controller with parameters $\theta_P$, $\mathcal{R}$ is the Retriever ($\mathcal{R}: \Sigma^* \to \R^d$), and $\mathcal{M}$ is the External Memory.
\end{customdefn}

\begin{figure}[t]
    \centering
    \includegraphics[width=0.85\textwidth]{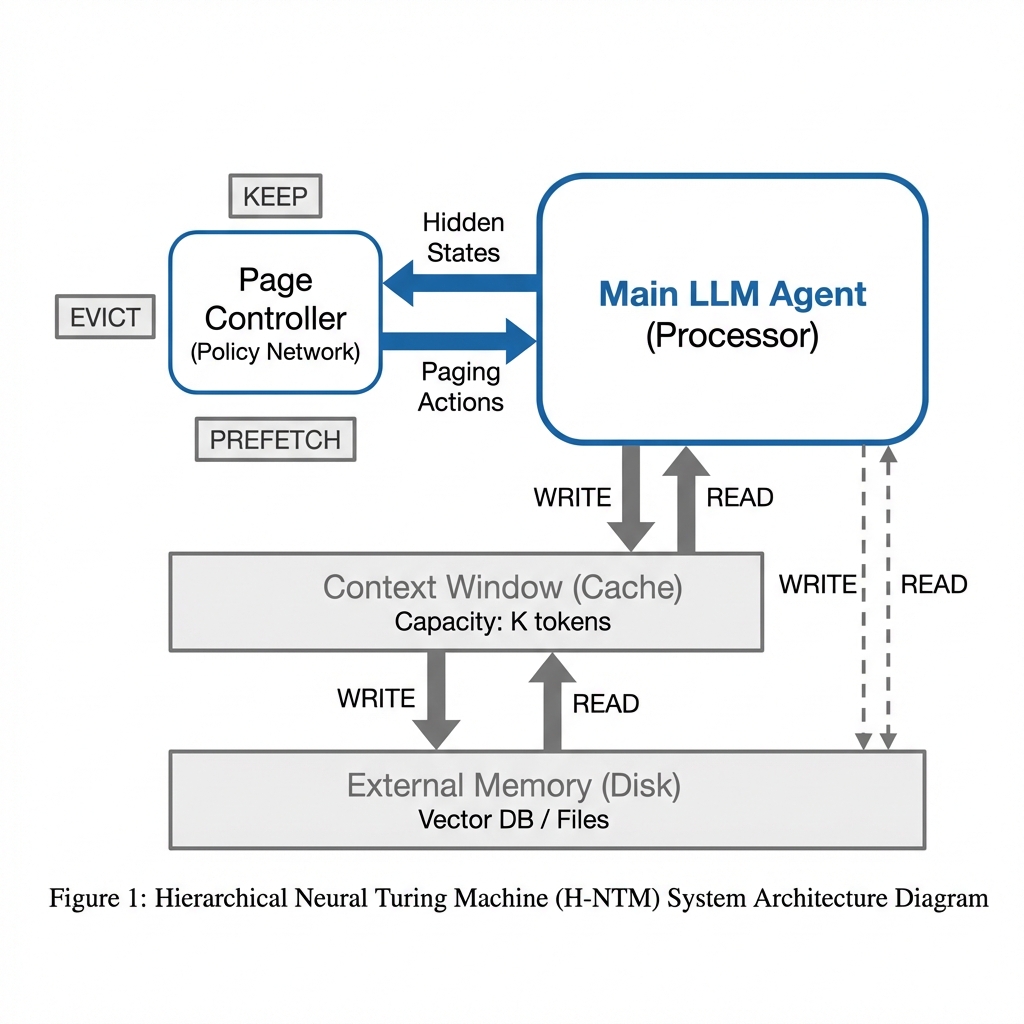}
    \caption{H-NTM System Architecture (schematic). The Main Agent (LLM) focuses on token generation. The Page Controller monitors activations and manages data flow between Context Window (Cache) and External Memory (Disk).}
    \label{fig:architecture}
\end{figure}

The key principle is \textit{information hiding}: the Main LLM operates as if it had a fixed-size context window, while the Page Controller transparently ensures that relevant information is available.

\subsection{Interface Design}

\begin{customdefn}{10}[Controller Interface]
The Page Controller's observation at time $t$ is $o_t = \mathrm{Encoder}(f_{\text{mode}}(\mathcal{L}, C_t, y_{<t}))$ where $f_{\text{mode}}$ depends on the operational mode:
\begin{center}
\begin{tabular}{lll}
\toprule
\textbf{Mode} & \textbf{Observable Information} & \textbf{Applicability} \\
\midrule
White-Box & Attention weights, hidden states, context & Open-source LLMs \\
Black-Box & Output tokens, context, logits & API-based LLMs \\
Gray-Box & Embeddings and outputs & Partially observable \\
\bottomrule
\end{tabular}
\end{center}
\end{customdefn}

\begin{customlem}{2}[Black-Box Sufficiency under Explicit Plans]
\label{lem:blackbox}
If there exists a mapping $g$ from output prefixes to optimal prefetch targets such that $P(g(y_{<t}) = b^*_{\text{prefetch}}) \geq 1 - \epsilon$ for all $t$, then a Black-Box controller implementing $g$ achieves expected page-fault rate within $\epsilon$ of the White-Box controller.
\end{customlem}

\begin{proof}
When $g$ is correct (probability $\geq 1-\epsilon$), the paging action matches the White-Box decision. Each incorrect prediction contributes at most one additional fault. Taking expectation yields an additive $\epsilon$ bound.
\end{proof}

\begin{remark}[Existence of $g$]
Lemma~\ref{lem:blackbox} is conditional. A mapping $g$ is plausible only when output prefixes contain explicit plan tokens or tool directives with stable identifiers. Agent frameworks like ReAct~\citep{yao2023react}, which produce structured action tokens (e.g., \texttt{Search[query]}), are natural candidates. For unconstrained generation, such a mapping is unlikely to exist.
\end{remark}

\subsection{Page Controller Architecture}

\begin{customdefn}{11}[Page Controller Network]
The Page Controller $\mathcal{P}_\theta: \mathcal{S} \to \Prob(\mathcal{A}^{|C_t|})$ parameterized by $\theta$, with mode-specific input encoding:

\textit{White-Box:}
$\mathcal{P}^{\text{WB}}_\theta(s_t) = \mathrm{Softmax}(W_{\text{out}} \cdot \mathrm{GELU}(W_2 \cdot \mathrm{LN}(W_1 \cdot [\Phi_C(C_t); \Phi_h(h_t); \Phi_A(A_t)])))$

\textit{Black-Box:}
$\mathcal{P}^{\text{BB}}_\theta(s_t) = \mathrm{Softmax}(W_{\text{out}} \cdot \mathrm{GELU}(W_2 \cdot \mathrm{LN}(W_1 \cdot [\Phi_C(C_t); \Phi_y(y_{<t}); \Phi_{\text{logits}}(\mathrm{logits}(y_t))])))$

\noindent where $\Phi_C$ is a mean-pooled block encoder, $\Phi_h$ a projected hidden state, $\Phi_A$ an attention pattern summary, $\Phi_y$ an output sequence encoder, and $\Phi_{\text{logits}}$ a logit distribution encoder.
\end{customdefn}

The action space for each block $b \in C_t$ is $\mathcal{A} = \{\mathrm{KEEP}, \mathrm{EVICT}, \mathrm{PREFETCH}(q)\}$ where $q \in \mathcal{Q}_r$.

\subsection{Semantic Value Estimation}

\begin{customdefn}{12}[Semantic Value Function]
\label{def:value}
The value of block $b$ in state $s_t$ is:
\begin{equation}
V_\theta(b, s_t) = \E_{\pi} \left[ \sum_{k=0}^{H-1} \gamma^k R_{\text{utility}}(b, s_{t+k}) \mid s_t \right]
\end{equation}
where $R_{\text{utility}}(b, s) = \mathrm{PredGain}(b; C \setminus b)$.
\end{customdefn}

The eviction policy selects $b_{\text{evict}} = \argmin_{b \in C_t} V_\theta(b, s_t)$.

\begin{figure}[t]
    \centering
    \includegraphics[width=0.85\textwidth]{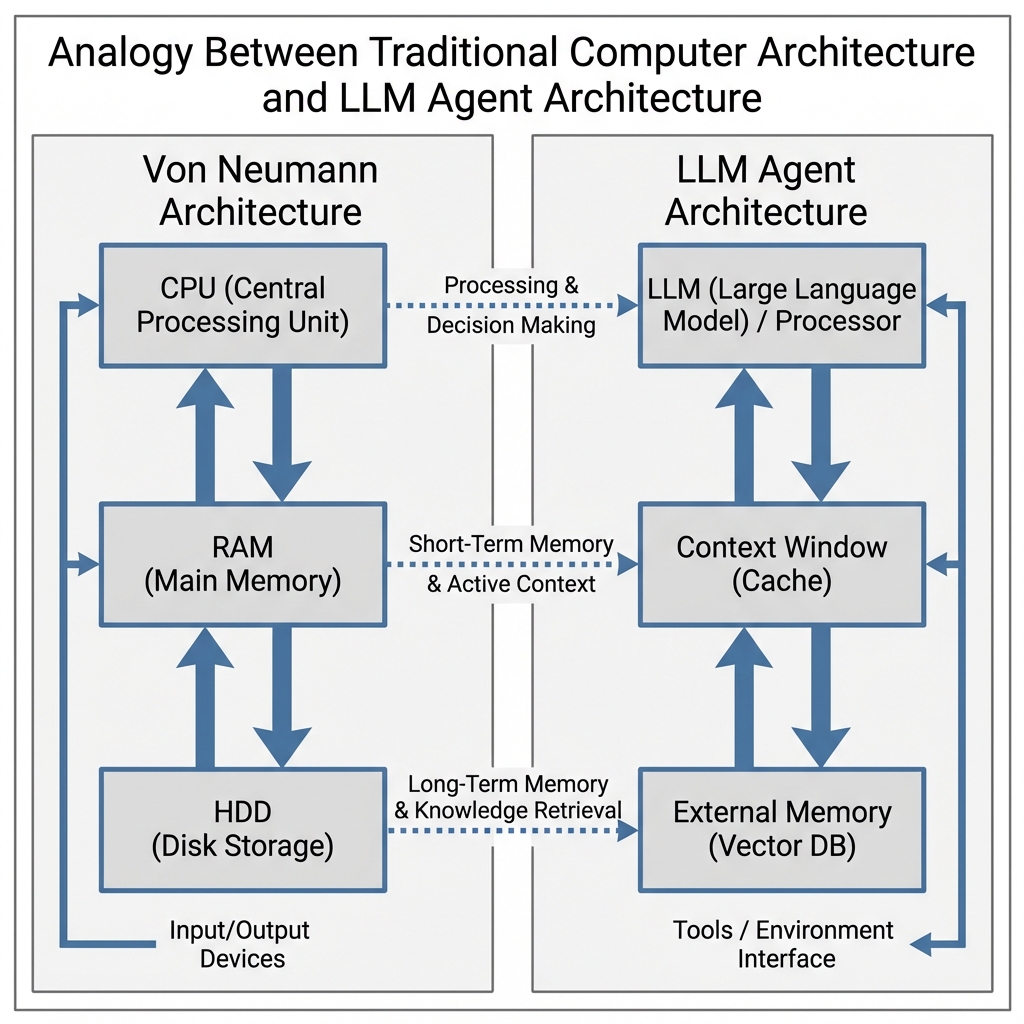}
    \caption{Context as Cache Hierarchy (schematic). The Context Window acts as L1/L2 Cache, requiring distinct management strategies from the massive External Knowledge Base.}
    \label{fig:analogy}
\end{figure}

\subsection{Predictive Prefetching via CoT Analysis}

\begin{customdefn}{13}[Intent Extraction]
Given partial output $y_{<t}$, the Intent Extractor produces $I_t = \mathrm{Extractor}(y_{<t}) = (e_{\text{next}}, e_{\text{entities}}, e_{\text{tools}})$, and the prefetch query is $q_t = W_q [e_{\text{next}}; e_{\text{entities}}; e_{\text{tools}}]$.
\end{customdefn}

\subsection{Training Objective and Algorithm}

\begin{customdefn}{14}[Paging CMDP]
\label{def:cmdp}
The Context Paging CMDP is $(\mathcal{S}, \mathcal{A}, P, R, g, \gamma)$ where:
\begin{itemize}
    \item State $\mathcal{S} = \{(C, E, h, y) : |C| \leq K_{\max}\}$
    \item Action $\mathcal{A} = \mathcal{A}_{\text{evict}} \times \mathcal{A}_{\text{fetch}}$ (per-block decisions $\times$ global prefetch)
    \item Transition $P: \mathcal{S} \times \mathcal{A} \to \Delta(\mathcal{S})$ (determined by frozen LLM and retriever)
    \item Reward $R(s,a) = \alpha \log P(y^* \mid C_t) - \lambda_{\text{evict}}\,n_{\text{evict}}(a) - \lambda_{\text{fetch}}\,n_{\text{fetch}}(a)$
    \item Constraint $g(s,a) = |C_t| - K_{\max} \leq 0$
\end{itemize}
\end{customdefn}

\begin{remark}[Constraint Handling]
The constraint can be enforced by action masking or via Lagrangian relaxation: $\max_\pi \min_{\mu \ge 0} \E[R(s,a)] - \mu\,\E[g(s,a)]$.
\end{remark}

\begin{algorithm}[t]
\caption{Neural Paging Training (PPO)}
\label{alg:training}
\begin{algorithmic}[1]
\STATE Initialize controller parameters $\theta$, critic $V_\theta$
\FOR{epoch $= 1$ to $N_{\text{epochs}}$}
    \STATE Roll out $\pi_{\theta}$ in LLM+retriever environment; collect $\tau = \{(s_t, a_t, r_t)\}_{t=1}^T$
    \STATE Compute GAE advantages $\hat{A}_t$ with $\lambda_{\text{GAE}} = 0.95$ and value targets $\hat{V}_t$
    \FOR{mini-batch $\mathcal{B} \subset \tau$, $N_{\text{inner}}$ epochs}
        \STATE Update $\theta$ by maximizing the clipped PPO objective:
        \STATE \quad $L(\theta) = \E_t[\min(r_t(\theta)\hat{A}_t, \mathrm{clip}(r_t(\theta), 1{-}\epsilon, 1{+}\epsilon)\hat{A}_t)] - c_v(V_\theta(s_t) - \hat{V}_t)^2 + c_e H(\pi_\theta)$
    \ENDFOR
\ENDFOR
\end{algorithmic}
\end{algorithm}

\textbf{Training regime.}
The Page Controller has discrete, finite actions (KEEP, EVICT, PREFETCH candidates), a state space bounded by $K$, and a reward signal at each paging decision. These properties make the problem more amenable to RL than typical continuous control. However, general convergence guarantees for neural network policies are unavailable due to non-convexity; \citet{schulman2017ppo} showed that PPO provides a lower bound on policy improvement at each step under the clipped surrogate, but formal convergence to a global optimum remains open.

\subsection{Handling Open-Ended Tasks}

\begin{customdefn}{15}[Uncertainty-Based Reward]
\label{def:uncertainty}
For open-ended generation, the prediction reward is:
\begin{equation}
R_{\text{pred}}(s_t) = -H(P(y_t \mid C_t)) = -\sum_y P(y \mid C_t) \log P(y \mid C_t).
\end{equation}
\end{customdefn}

\begin{remark}[Reward Exploitation]
The uncertainty-based reward may be exploited by policies that always prefetch ``common'' information. To mitigate this, we add a novelty-weighted exploration bonus:
\begin{equation}
R_{\text{explore}}(b) = \frac{\eta}{\sqrt{N_{\text{fetch}}(b) + 1}}
\end{equation}
where $N_{\text{fetch}}(b)$ counts prior fetches of block $b$ and $\eta > 0$ controls exploration strength. This encourages the controller to prefetch diverse, task-relevant blocks rather than repeatedly fetching familiar ones.
\end{remark}

\begin{figure}[t]
    \centering
    \includegraphics[width=0.85\textwidth]{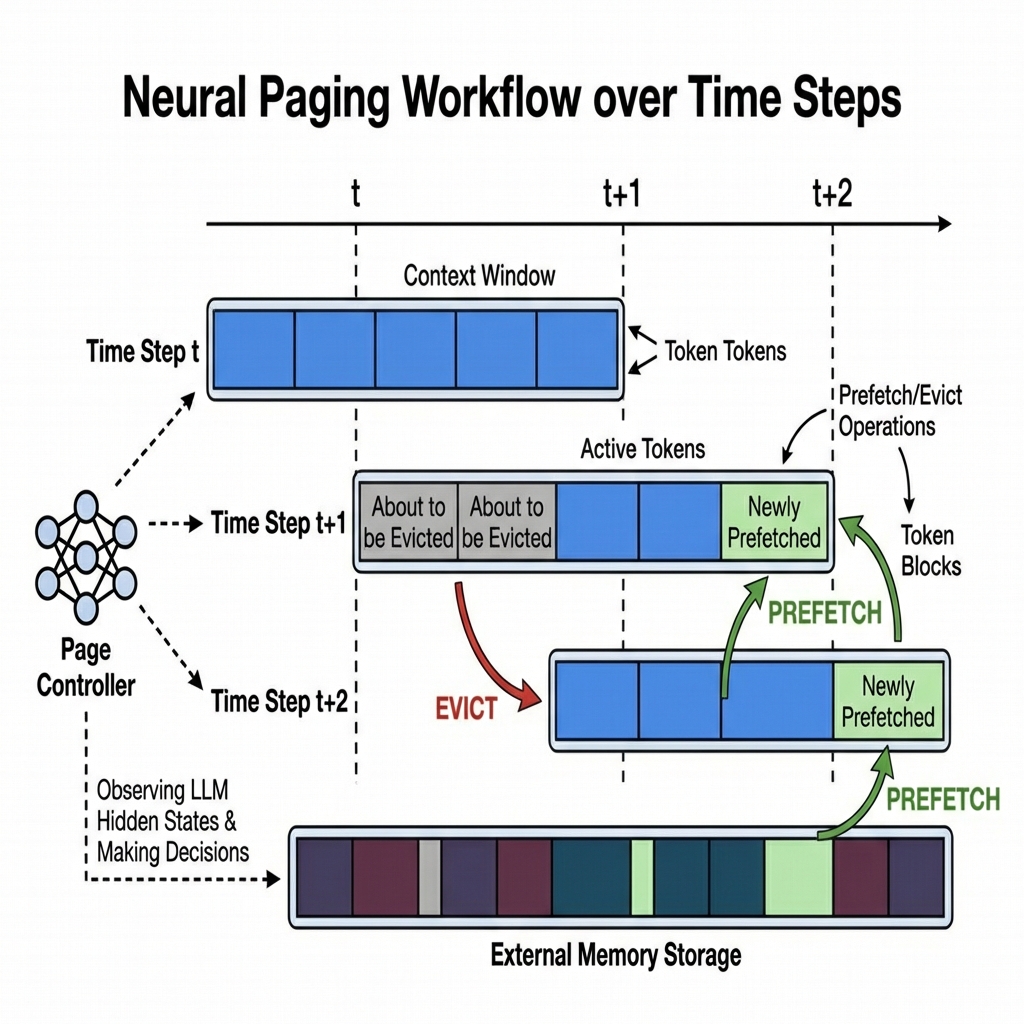}
    \caption{Neural Paging Workflow (schematic). As reasoning progresses from $t$ to $t{+}2$, old blocks are evicted and new blocks are prefetched, maintaining a dynamic window of semantic relevance.}
    \label{fig:workflow}
\end{figure}

% ====================================================================
\section{Theoretical Analysis}
\label{sec:analysis}
% ====================================================================

\subsection{Competitive Ratio Analysis}

\begin{customthm}{3}[Competitive Ratio Lower Bound, Classical Model]
\label{thm:lower}
Under Assumption~1, no deterministic online paging algorithm can achieve a competitive ratio better than $K_b$ against the optimal offline policy.
\end{customthm}

\begin{proof}
The adversary constructs a request sequence over $K_b+1$ distinct blocks, cycling to force any online algorithm to repeatedly evict the block that will be requested next, while the optimal offline algorithm retains the correct $K_b$ blocks. This is the classical paging lower bound~\citep{sleator1985paging}; under Assumption~1, the induced block accesses form an exogenous stream and the bound applies unchanged.
\end{proof}

\begin{customthm}{4}[Competitive Ratio under Bounded Sensitivity]
\label{thm:robustness}
Suppose access sequences satisfy $\beta$-bounded sensitivity (Definition~\ref{def:sensitivity}). If an online paging algorithm $A$ is $c$-competitive on exogenous sequences and the offline optimum also satisfies the sensitivity bound, then for any policy-induced sequence $r^\pi$:
\begin{equation}
F_A(r^\pi) \leq c\,F_{\mathrm{opt}}(r^\pi) + (c+1)(K_b + 1)\,\beta T.
\label{eq:thm4}
\end{equation}
\end{customthm}

\begin{proof}
Fix a reference policy $\pi_0$ inducing sequence $r^0 := r^{\pi_0}$. By Definition~\ref{def:sensitivity}, $d_H(r^\pi, r^0) \leq \beta T$. By Lemma~\ref{lem:sensitivity}:
\begin{align}
|F_A(r^\pi) - F_A(r^0)| &\leq (K_b+1)\,\beta T, \label{eq:sens_a}\\
|F_{\mathrm{opt}}(r^\pi) - F_{\mathrm{opt}}(r^0)| &\leq (K_b+1)\,\beta T. \label{eq:sens_opt}
\end{align}
Since $A$ is $c$-competitive on the fixed sequence $r^0$: $F_A(r^0) \leq c\,F_{\mathrm{opt}}(r^0)$. Chaining:
\begin{align*}
F_A(r^\pi) &\leq F_A(r^0) + (K_b{+}1)\beta T \\
&\leq c\,F_{\mathrm{opt}}(r^0) + (K_b{+}1)\beta T \\
&\leq c\,(F_{\mathrm{opt}}(r^\pi) + (K_b{+}1)\beta T) + (K_b{+}1)\beta T \\
&= c\,F_{\mathrm{opt}}(r^\pi) + (c{+}1)(K_b{+}1)\beta T. \qedhere
\end{align*}
\end{proof}

\begin{remark}
When $\beta = 0$ (Assumption~1), Theorem~\ref{thm:robustness} recovers the classical $c$-competitive bound. The additive term $(c{+}1)(K_b{+}1)\beta T$ quantifies the ``price of non-exogeneity.'' For LRU ($c = K_b$), the bound becomes $K_b \cdot F_{\mathrm{opt}} + (K_b{+}1)^2 \beta T$. On structured traces where $\beta \ll 1$ (see Section~\ref{sec:beta_estimation}), the additive penalty is small relative to $T$.
\end{remark}

\begin{customprop}{6}[Bound under Bounded Error Impact]
\label{prop:error}
If the Page Controller predicts the optimal eviction target with accuracy $p$ and each incorrect eviction increases faults by at most $D_f$, then:
\begin{equation}
\E[F_{\text{learned}}] \leq F_{\mathrm{opt}} + (1-p)\,D_f\,T.
\end{equation}
\end{customprop}

\begin{proof}
At each of $T$ eviction points, a correct prediction (probability $p$) matches Belady; an incorrect one (probability $1{-}p$) adds at most $D_f$ faults. Linearity of expectation gives the bound. The bounded-error and independence assumptions are strong; correlated errors and cascade effects (cf.\ Lemma~\ref{lem:sensitivity}) may amplify $D_f$ in practice.
\end{proof}

\subsection{Policy Comparison (Qualitative)}

\begin{customdefn}{16}[Policy Classes]
We define: $\Pi_H = \{\text{LRU}, \text{LFU}, \text{FIFO}, \text{Random}\}$ (heuristic), $\Pi_S = \{\text{Fixed}\}$ (static), $\Pi_L = \{\text{Neural Paging}\}$ (learned).
\end{customdefn}

Under Assumption~1, if a learned policy matches Belady's eviction decision on a larger fraction of steps than a heuristic, its expected fault count is lower by Proposition~\ref{prop:error}. Determining whether this holds is an empirical question. The synthetic experiments in Section~\ref{sec:experiments} show that on structured traces, the gap between heuristics and Belady is substantial (competitive ratio $\approx 1.9$ vs.\ worst-case bound $K_b = 8$), indicating significant room for learned policies.

\subsection{Complexity-Theoretic Considerations}

\begin{customdefn}{17}[Finite-Horizon Optimal Paging Problem]
Given a finite MDP $(\mathcal{S}, \mathcal{A}, P, R, \gamma)$ with horizon $H$, compute the optimal first action $a_0^*$ maximizing $\E[\sum_{k=0}^{H-1} \gamma^k R(s_k, a_k)]$.
\end{customdefn}

\begin{customprop}{4}[PSPACE-Membership]
The Finite-Horizon Optimal Paging Problem is in PSPACE when $|\mathcal{S}|$, $|\mathcal{A}|$, and $H$ are polynomial in the input size.
\end{customprop}

\begin{proof}[Proof Sketch]
Backward induction computes $V_t(s) = \max_a \E[R(s,a) + \gamma V_{t+1}(s')]$ for all $s$, using $O(|\mathcal{S}| \cdot |\mathcal{A}|)$ space per layer and overwriting previous layers. PSPACE-hardness for general transitions remains open.
\end{proof}

\subsection{Expressiveness of the Policy Class}

\begin{customprop}{5}[Approximation of Smooth Policies]
\label{prop:approx}
For any Lipschitz-continuous policy $\pi^*: \mathcal{S} \to \Delta(\mathcal{A})$ and any $\epsilon > 0$, there exists a neural network $\mathcal{P}_\theta$ with $\sup_{s \in \mathcal{S}'} \|\mathcal{P}_\theta(s) - \pi^*(s)\|_1 < \epsilon$ on any compact $\mathcal{S}' \subseteq \mathcal{S}$.
\end{customprop}

\begin{proof}
By the Universal Approximation Theorem~\citep{cybenko1989, hornik1991}. However, optimal paging policies for discrete decisions may have sharp boundaries that violate Lipschitz continuity. The approximation applies only to the restricted class of smooth policies; in practice, neural policies learn smooth surrogates that may incur non-negligible $\epsilon$.
\end{proof}

\subsection{Observability and POMDP Structure}
\label{sec:pomdp}

The MDP formulation in Definition~\ref{def:cmdp} assumes the controller observes the full state $s_t = (C_t, E_t, h_t, y_t)$. In practice, the observability depends on the interface mode (Definition~10):

\textbf{White-Box.} The controller accesses attention weights and hidden states, which are functions of $C_t$ and the model's internal computation. Under mild assumptions (e.g., that attention patterns summarize the information needed for paging decisions), the Markov property approximately holds.

\textbf{Black-Box.} The controller observes only $(y_{<t}, C_t, \mathrm{logits}(y_t))$. The hidden state $h_t$---which encodes the model's internal reasoning and may predict future information needs---is unobserved. This renders the problem a POMDP. Consequently, theoretical results that rely on the full state (Proposition~4, the CMDP formulation) apply to the White-Box setting but may not transfer directly to Black-Box. In the Black-Box setting, the controller must maintain a belief state or use history-dependent policies (e.g., recurrent architectures) to compensate.

\textbf{Practical impact.} The gap between White-Box and Black-Box depends on how much information about future needs is encoded in hidden states vs.\ externalized in output tokens. For agents using structured reasoning formats (e.g., ReAct~\citep{yao2023react} with explicit tool-call tokens), the gap may be small; for free-form generation, it can be significant. Lemma~\ref{lem:blackbox} formalizes the condition under which the gap vanishes.

\subsection{Summary of Theoretical Results}

\begin{table}[H]
\centering
\caption{Summary of Theoretical Results}
\label{tab:results}
\begin{tabular}{lll}
\toprule
\textbf{Result} & \textbf{Statement} & \textbf{Status} \\
\midrule
Thm~1 & Turing completeness with $O(T_{\mathrm{TM}})$ simulation & Classical; explicit conditions \\
Thm~2 & Complexity $O(N \cdot K^2)$ & Holds for any fixed-$K$ method \\
Thm~3 & Competitive ratio $\geq K_b$ & Classical lower bound \\
\textbf{Thm~4} & \textbf{Robustness under $\beta$-sensitivity} & \textbf{New; validated synthetically} \\
Prop~5 & Neural approximation of smooth policies & Standard UAT application \\
\bottomrule
\end{tabular}
\end{table}

% ====================================================================
\section{Synthetic Validation}
\label{sec:experiments}
% ====================================================================

We validate the theoretical bounds on synthetic paging traces with controlled parameters. The goals are to: (i)~confirm that Theorem~\ref{thm:robustness} holds empirically, (ii)~characterize the tightness of the bounds, and (iii)~quantify the gap between heuristic policies and the offline optimum on structured traces.

\subsection{Experimental Setup}
\label{sec:exp_setup}

\textbf{Trace generation.}
We generate access traces from a non-stationary Zipf distribution over $M = 64$ blocks with exponent $\alpha = 1.2$ and a working set of size~16 that shifts every 500 steps. This models the typical access pattern of a long-horizon agent: a ``hot'' set of frequently accessed blocks with periodic phase transitions. Each trace has length $T = 5{,}000$.

\textbf{Controlled sensitivity.}
To test Theorem~\ref{thm:robustness}, we create pairs of traces with controlled Hamming distance. Given a base trace $r^0$, we produce a perturbed trace $r^\beta$ by flipping exactly $\lfloor \beta T \rfloor$ randomly chosen positions to uniformly random blocks, yielding $d_H(r^0, r^\beta) = \lfloor \beta T \rfloor$. We vary $\beta \in \{0, 0.02, 0.05, 0.1, 0.15, 0.2, 0.3, 0.4, 0.5\}$.

\textbf{Algorithms.}
We compare five paging algorithms: Belady (optimal offline), LRU, LFU, FIFO, and Random eviction. All operate with cache size $K_b \in \{2, 4, 6, 8, 10, 12, 16\}$.

\textbf{Metrics.}
Fault rate (faults$/T$), competitive ratio ($F_A / F_{\mathrm{opt}}$), and fault stability ($|F_A(r^\beta) - F_A(r^0)|$). All results are averaged over 10 random seeds with standard deviations reported.

\begin{figure}[t]
    \centering
    \includegraphics[width=\textwidth]{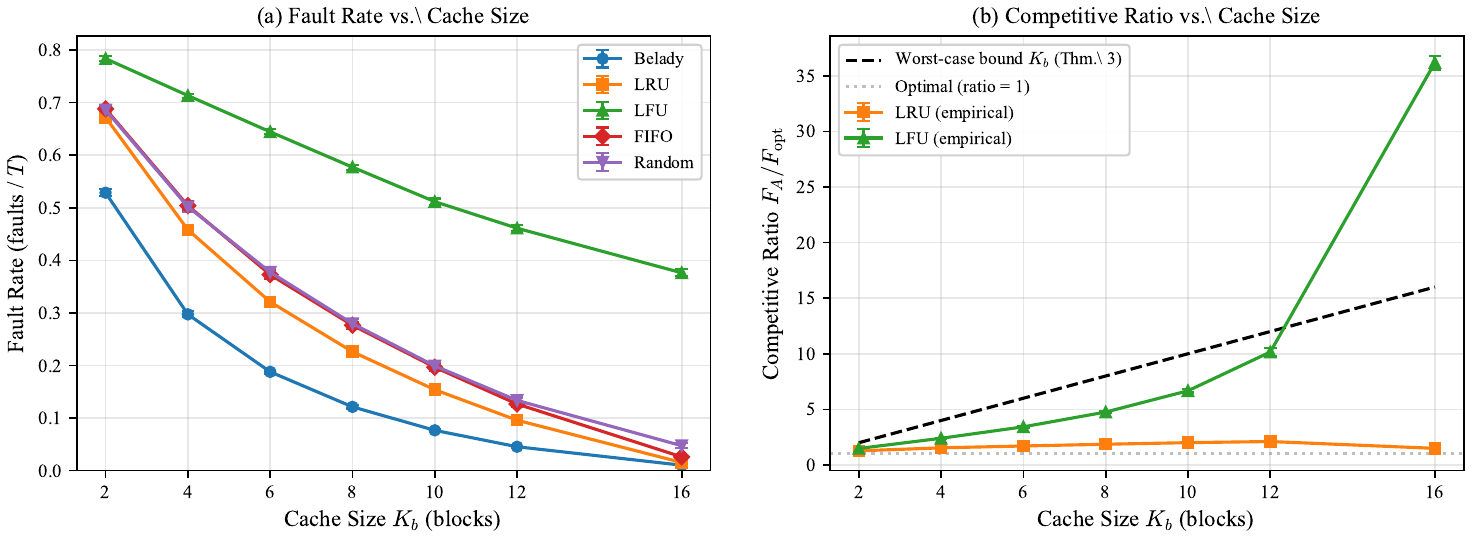}
    \caption{(a)~Fault rate vs.\ cache size on Zipf traces ($M{=}64$, $T{=}5{,}000$). Belady is optimal; LRU is the best online heuristic. LFU suffers from frequency poisoning on shifting working sets. (b)~Empirical competitive ratio vs.\ worst-case bound $K_b$ (Theorem~3). On structured traces, online algorithms perform far better than worst-case, with LRU at ${\approx}1.9\times$ optimal. Error bars: $\pm 1$ s.d.\ over 10 seeds.}
    \label{fig:scaling}
\end{figure}

\subsection{Fault Rate Scaling (Figure~\ref{fig:scaling})}

Figure~\ref{fig:scaling}(a) shows fault rates as a function of cache size. Key observations:
\begin{itemize}
    \item Belady achieves the lowest fault rate at all cache sizes, as expected. At $K_b = 8$, its fault rate is $0.121 \pm 0.003$.
    \item LRU is the best online heuristic ($0.226 \pm 0.007$ at $K_b = 8$), followed by FIFO ($0.276$) and Random ($0.280$). LFU performs poorly ($0.577$) because it retains historically frequent blocks from past working sets rather than adapting to shifts.
    \item The gap between Belady and LRU narrows as $K_b$ increases, consistent with the working set model: when the cache can hold the entire working set, all reasonable policies converge.
\end{itemize}

Figure~\ref{fig:scaling}(b) shows empirical competitive ratios. LRU achieves a ratio of $1.86 \pm 0.04$ at $K_b = 8$---far below the worst-case bound of $K_b = 8$ from Theorem~\ref{thm:lower}. This gap between empirical and worst-case performance demonstrates that structured access patterns (Zipf with locality) are much more benign than adversarial sequences, and motivates learning policies that exploit this structure.

\begin{figure}[t]
    \centering
    \includegraphics[width=\textwidth]{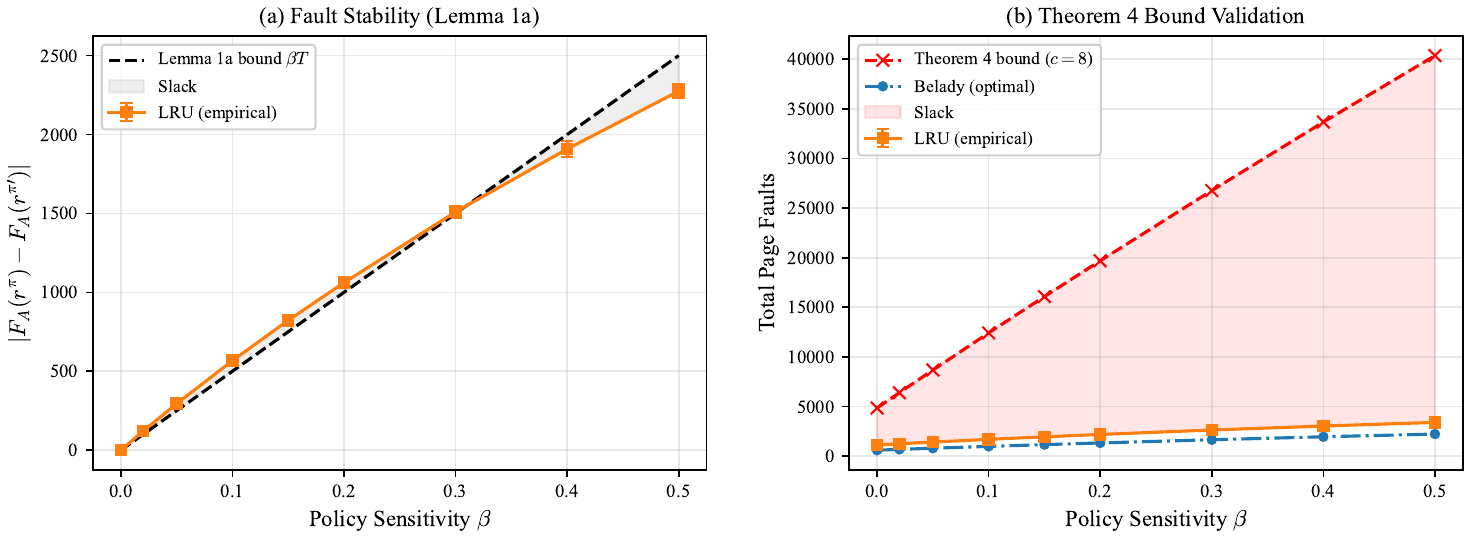}
    \caption{Theorem~4 validation ($K_b{=}8$, $T{=}5{,}000$). (a)~Fault stability: empirical $|F_A(r^\beta) - F_A(r^0)|$ vs.\ reference line $\beta T$. The empirical values slightly exceed $\beta T$ at small $\beta$ (cascade effect), but remain well within the corrected $(K_b{+}1)\beta T$ bound. (b)~Theorem~4 bound (red dashed) vs.\ actual LRU faults (orange). The bound holds with large slack, indicating room for tighter instance-dependent analysis. Error bars: $\pm 1$ s.d.\ over 10 seeds.}
    \label{fig:theorem4}
\end{figure}

\subsection{Theorem~4 Validation (Figure~\ref{fig:theorem4})}

Figure~\ref{fig:theorem4}(a) validates the fault-sensitivity bound from Lemma~\ref{lem:sensitivity}. The empirical fault difference $|F_{\text{LRU}}(r^\beta) - F_{\text{LRU}}(r^0)|$ grows linearly with $\beta$, closely tracking the reference line $\beta T$. At small~$\beta$ ($\leq 0.15$), the empirical values slightly exceed $\beta T$ (by a factor of ${\approx}1.13$), confirming that the cascade effect described in Lemma~\ref{lem:sensitivity} is real but mild. The corrected bound $(K_b{+}1)\beta T = 9\,\beta T$ holds with large margin at all tested values.

Figure~\ref{fig:theorem4}(b) validates Theorem~\ref{thm:robustness}. The Theorem~4 upper bound $c \cdot F_{\mathrm{opt}} + (c{+}1)(K_b{+}1)\beta T$ (computed with $c = K_b = 8$) is satisfied at all $\beta$ values, with substantial slack. This slack arises from two sources: (i)~the empirical competitive ratio is ${\approx}1.9$ vs.\ worst-case $c = 8$, and (ii)~the cascade factor is ${\approx}1.13$ vs.\ worst-case $K_b{+}1 = 9$. Both sources of conservatism can be reduced by deriving instance-dependent bounds---an avenue for future work.

\subsection{Summary of Experimental Findings}

\begin{enumerate}
    \item \textbf{Bounds are correct:} Theorem~4's bound is satisfied for all tested $\beta$ values.
    \item \textbf{Cascade is mild:} The empirical cascade factor (${\approx}1.13$) is far below the worst-case $K_b{+}1 = 9$, suggesting that the $(K_b{+}1)$ factor in Lemma~\ref{lem:sensitivity} is pessimistic for structured traces.
    \item \textbf{Large gap between heuristics and optimum:} LRU's competitive ratio of ${\approx}1.9$ (vs.\ worst-case 8) shows that structured access patterns offer significant room for improvement---the core motivation for learning paging policies.
    \item \textbf{LFU is fragile:} On non-stationary traces, LFU can be much worse than LRU (competitive ratio ${\approx}4.8$ vs.\ ${\approx}1.9$), highlighting the danger of using heuristics mismatched to the access pattern. This further motivates adaptive, learned policies.
\end{enumerate}

% ====================================================================
\section{Discussion}
\label{sec:discussion}
% ====================================================================

\subsection{Scope and Applicability}
Neural Paging is designed for stateful, long-horizon agent tasks where information needs evolve over time:

\begin{center}
\begin{tabular}{lll}
\toprule
\textbf{Scenario} & \textbf{Suitability} & \textbf{Reason} \\
\midrule
Multi-step reasoning & High & Working set changes predictably ($\beta \leq 0.05$) \\
Streaming data & High & Continuous context turnover \\
Interactive dialogue & Medium & Context grows but rarely shrinks \\
Single-pass QA & Low & No eviction needed \\
\bottomrule
\end{tabular}
\end{center}

\subsection{Limitations}

\textit{No end-to-end evaluation.} The synthetic experiments validate theoretical bounds on abstract paging traces. They do not demonstrate that Neural Paging improves task accuracy when integrated with a real LLM. A minimal validation path is: (1)~synthetic traces (this paper), (2)~retrieval noise sweeps with controlled $\rho$, (3)~end-to-end agent tasks with frozen LLM measuring token cost, latency, and quality. Steps 2--3 are essential next steps.

\textit{Training complexity.} The Page Controller requires task-specific training. Transfer across domains remains an open problem.

\textit{Prediction uncertainty.} Future information needs are inherently uncertain; incorrect value estimates may cause suboptimal evictions.

\textit{Cold start.} At episode start, the controller has limited information, causing early page faults before adaptation.

\textit{Retrieval dependencies.} Neural Paging assumes reliable retrieval; if $\phi$ fails, faults cannot be resolved.

\textit{Theoretical conservatism.} The $(K_b{+}1)$ factor in Lemma~\ref{lem:sensitivity} and the worst-case competitive ratio $c = K_b$ in Theorem~\ref{thm:robustness} make the bounds conservative. Instance-dependent analysis (e.g., using the actual competitive ratio on specific trace distributions) could yield much tighter guarantees.

\subsection{Connections to Computability Theory}

The H-NTM maps cleanly onto the components of a Turing machine: the LLM corresponds to finite control, the Context Window to the tape segment under the head, and External Memory to the full tape. The Page Controller implements ``tape head movement'' by managing which tape cells are in context. Neural Paging can thus be viewed as resource-bounded computation where the bound is the context window size~$K$ rather than time or space.

\subsection{Multi-Agent Context Partitioning (Sketch)}

\begin{customdefn}{18}[Multi-Agent H-NTM]
A multi-agent system $\{\mathcal{H}_i\}_{i=1}^n$ with shared external memory $E_{\text{shared}}$. Each agent $i$ has context $C_t^i$ and policy $\pi_i$, with $\sum_{i=1}^n |C_t^i| \leq K_{\text{shared}}$. The joint policy may be decentralized or coordinated.
\end{customdefn}

This setting raises questions about cache contention, dynamic capacity allocation, and cooperative objectives. We include it as a concrete extension but do not pursue formal results.

\subsection{Open Problems}
\begin{enumerate}
    \item \textit{Instance-dependent competitive bounds}: Derive tighter bounds for specific trace distributions (e.g., Zipf), improving on the worst-case $K_b$.
    \item \textit{Tight cascade analysis}: Prove that the empirical cascade factor $O(1)$ (rather than $K_b{+}1$) holds for traces with locality, and characterize the locality conditions formally.
    \item \textit{Hierarchical paging}: Extend to multi-level caches (L1/L2/L3 analog).
    \item \textit{Adaptive block sizes}: Allow variable block sizes based on semantic coherence.
    \item \textit{Joint optimization}: End-to-end training of LLM and Page Controller with architectural modifications.
    \item \textit{Predictable access and improved guarantees}: Characterize tasks where future access is predictable from context and derive competitive ratios below $K_b$.
\end{enumerate}

% ====================================================================
\section{Conclusion}
\label{sec:conclusion}
% ====================================================================

We presented the Hierarchical Neural Turing Machine and the Neural Paging framework as a principled approach to context management for long-horizon AI agents. The primary contributions are: (1)~the formal Context Paging Problem with a bounded-sensitivity access model; (2)~the H-NTM architecture that decouples reasoning from memory management; (3)~Theorem~4, a new robustness bound quantifying competitive-ratio degradation under policy-dependent access; and (4)~synthetic experiments confirming the bounds and revealing substantial slack that motivates learned paging policies.

Key findings from the synthetic validation---that structured access patterns yield empirical competitive ratios far below worst-case bounds, and that the cascade effect from access perturbation is mild---together provide both the theoretical justification and the empirical motivation for deploying learned context management in real agent systems. End-to-end evaluation with frozen LLMs on long-horizon tasks is the natural and essential next step.

% ====================================================================
\appendix
\section{Proof Details}
\label{app:proofs}

\subsection{Proof of Theorem~1}
\label{app:thm1}
We sketch a concrete simulation. Represent the TM tape as fixed-size blocks of $B$ tokens stored in external memory, each tagged with its address. By Assumption~3, the controller can fetch any block by address in $O(1)$ queries.

At each TM step, the controller fetches the block containing the head position (and one adjacent block if the head crosses a boundary), applies the transition, and writes back. The context window holds $O(B)$ tokens. Each step costs $O(B^2)$ attention and $O(1)$ retrieval. Over $T_{\mathrm{TM}}(n)$ steps: $O(T_{\mathrm{TM}}(n) \cdot B^2) = O(T_{\mathrm{TM}}(n))$ for constant $B$.

\subsection{Proof of Theorem~2}
At each of $N$ generation steps, self-attention costs $O(K^2)$, yielding $O(NK^2)$. Paging decisions occur every $B$ tokens; retrieval via ANN costs $O(\log M)$ per query for $K_b$ candidates, totaling $O(N K_b \log M / B)$. Policy inference costs $O(K_b^2)$ per decision, totaling $O(N K_b^2 / B)$.

\subsection{Proof of Theorem~3}
Under Assumption~1, the classical adversarial construction over $K_b{+}1$ pages applies unchanged~\citep{sleator1985paging}.

\subsection{Proof of Theorem~4}
Full proof in Section~\ref{sec:analysis}. The key steps: (1)~apply Lemma~\ref{lem:sensitivity} to both $A$ and OPT with sensitivity $\beta$; (2)~use the $c$-competitive guarantee on the fixed reference sequence; (3)~chain the inequalities.

\subsection{Proof of Proposition~4}
Backward induction: $V_t(s) = \max_a \E[R(s,a) + \gamma V_{t+1}(s')]$. Working space per layer: $O(|\mathcal{S}| \cdot |\mathcal{A}|)$. Layers can be overwritten. Total space: $O(|\mathcal{S}| \cdot |\mathcal{A}|)$.

\subsection{Proof of Proposition~6}
At each of $T$ eviction points, correct prediction (prob $p$) matches Belady; error (prob $1{-}p$) adds $\leq D_f$ faults. Linearity of expectation: $\E[F_{\text{learned}}] \leq F_{\mathrm{opt}} + (1{-}p) D_f T$.

\section{Experimental Details}
\label{app:experiments}

\textbf{Trace generation.} Zipf distribution with exponent $\alpha = 1.2$, working set size 16, shift interval 500, total blocks $M = 64$, trace length $T = 5{,}000$. Ten random seeds (42--51) for all experiments.

\textbf{Perturbation.} For each $\beta$, exactly $\lfloor \beta T \rfloor$ positions are selected uniformly at random and replaced with uniformly random blocks (distinct from the original).

\textbf{Algorithms.} All algorithms implemented in Python with $O(T \log T)$ Belady using precomputed next-use indices. Full source code: \texttt{synthetic\_validation.py}.

% ====================================================================
\bibliographystyle{plainnat}
\bibliography{references}

@article{graves2014ntm,
  title   = {Neural Turing Machines},
  author  = {Graves, Alex and Wayne, Greg and Danihelka, Ivo},
  journal = {arXiv preprint arXiv:1410.5401},
  year    = {2014}
}

@article{graves2016dnc,
  title   = {Hybrid Computing Using a Neural Network with Dynamic External Memory},
  author  = {Graves, Alex and Wayne, Greg and Reynolds, Malcolm and Harley, Tim
             and Danihelka, Ivo and Grabska-Barwi{\'n}ska, Agnieszka and
             Colmenarejo, Sergio G{\'o}mez and Grefenstette, Edward and
             Ramalho, Tiago and Agapiou, John and others},
  journal = {Nature},
  volume  = {538},
  number  = {7626},
  pages   = {471--476},
  year    = {2016},
  publisher = {Nature Publishing Group}
}

@article{weston2014memory,
  title   = {Memory Networks},
  author  = {Weston, Jason and Chopra, Sumit and Bordes, Antoine},
  journal = {arXiv preprint arXiv:1410.3916},
  year    = {2014}
}

@article{schuurmans2023universal,
  title   = {Memory Augmented Large Language Models are Computationally Universal},
  author  = {Schuurmans, Dale},
  journal = {arXiv preprint arXiv:2301.04589},
  year    = {2023}
}

@article{dao2022flashattention,
  title   = {{FlashAttention}: Fast and Memory-Efficient Exact Attention
             with {IO}-Awareness},
  author  = {Dao, Tri and Fu, Dan and Ermon, Stefano and Rudra, Atri
             and R{\'e}, Christopher},
  journal = {Advances in Neural Information Processing Systems},
  volume  = {35},
  year    = {2022}
}

@article{dai2019transformerxl,
  title   = {{Transformer-XL}: Attentive Language Models Beyond a
             Fixed-Length Context},
  author  = {Dai, Zihang and Yang, Zhilin and Yang, Yiming and
             Carbonell, Jaime and Le, Quoc V and Salakhutdinov, Ruslan},
  journal = {arXiv preprint arXiv:1901.02860},
  year    = {2019}
}

@article{rae2020compressive,
  title   = {Compressive Transformers for Long-Range Sequence Modelling},
  author  = {Rae, Jack W and Potapenko, Anna and Jayakumar, Siddhant M
             and Hillier, Chloe and Lillicrap, Timothy P},
  journal = {arXiv preprint arXiv:1911.05507},
  year    = {2020}
}

@article{chen2023extending,
  title   = {Extending Context Window of Large Language Models via
             Positional Interpolation},
  author  = {Chen, Shouyuan and Wong, Sherman and Chen, Liangjian
             and Tian, Yuandong},
  journal = {arXiv preprint arXiv:2306.15595},
  year    = {2023}
}

@article{beltagy2020longformer,
  title   = {Longformer: The Long-Document Transformer},
  author  = {Beltagy, Iz and Peters, Matthew E and Cohan, Arman},
  journal = {arXiv preprint arXiv:2004.05150},
  year    = {2020}
}

@article{zaheer2020bigbird,
  title   = {{Big Bird}: Transformers for Longer Sequences},
  author  = {Zaheer, Manzil and Guruganesh, Guru and Dubey, Kumar Avinava
             and Ainslie, Joshua and Alberti, Chris and Ontanon, Santiago
             and Pham, Philip and Ravula, Anirudh and Wang, Qifan and
             Yang, Li and Ahmed, Amr},
  journal = {Advances in Neural Information Processing Systems},
  volume  = {33},
  year    = {2020}
}

@article{liu2023ring,
  title   = {Ring Attention with Blockwise Transformers for
             Near-Infinite Context},
  author  = {Liu, Hao and Zaharia, Matei and Abbeel, Pieter},
  journal = {arXiv preprint arXiv:2310.01889},
  year    = {2023}
}

@article{munkhdalai2024infini,
  title   = {Leave No Context Behind: Efficient Infinite Context
             Transformers with {Infini-attention}},
  author  = {Munkhdalai, Tsendsuren and Faruqui, Manaal and Gopal, Siddharth},
  journal = {arXiv preprint arXiv:2404.07143},
  year    = {2024}
}

@inproceedings{lewis2020rag,
  title     = {Retrieval-Augmented Generation for Knowledge-Intensive
               {NLP} Tasks},
  author    = {Lewis, Patrick and Perez, Ethan and Piktus, Aleksandra
               and Petroni, Fabio and Karpukhin, Vladimir and Goyal, Naman
               and K{\"u}ttler, Heinrich and Lewis, Mike and Yih, Wen-tau
               and Rockt{\"a}schel, Tim and Riedel, Sebastian and Kiela, Douwe},
  booktitle = {Advances in Neural Information Processing Systems},
  volume    = {33},
  year      = {2020}
}

@article{borgeaud2022retro,
  title   = {Improving Language Models by Retrieving from Trillions of Tokens},
  author  = {Borgeaud, Sebastian and Mensch, Arthur and Hoffmann, Jordan
             and Cai, Trevor and Rutherford, Eliza and Millican, Katie
             and van den Driessche, George Bm and Lespiau, Jean-Baptiste
             and Damoc, Bogdan and Clark, Aidan and others},
  journal = {International Conference on Machine Learning},
  year    = {2022}
}

@article{asai2023selfrag,
  title   = {{Self-RAG}: Learning to Retrieve, Generate, and Critique
             through Self-Reflection},
  author  = {Asai, Akari and Wu, Zeqiu and Wang, Yizhong and Sil, Avirup
             and Hajishirzi, Hannaneh},
  journal = {arXiv preprint arXiv:2310.11511},
  year    = {2023}
}

@article{yao2023react,
  title   = {{ReAct}: Synergizing Reasoning and Acting in Language Models},
  author  = {Yao, Shunyu and Zhao, Jeffrey and Yu, Dian and Du, Nan
             and Shafran, Izhak and Narasimhan, Karthik and Cao, Yuan},
  journal = {International Conference on Learning Representations},
  year    = {2023}
}

@article{schick2023toolformer,
  title   = {Toolformer: Language Models Can Teach Themselves to Use Tools},
  author  = {Schick, Timo and Dwivedi-Yu, Jane and Dess{\`i}, Roberto
             and Raileanu, Roberta and Lomeli, Maria and Hambro, Eric
             and Zettlemoyer, Luke and Cancedda, Nicola and Scialom, Thomas},
  journal = {Advances in Neural Information Processing Systems},
  volume  = {36},
  year    = {2023}
}

@article{packer2023memgpt,
  title   = {{MemGPT}: Towards {LLMs} as Operating Systems},
  author  = {Packer, Charles and Wooders, Sarah and Lin, Kevin and
             Fang, Vivian and Patil, Shishir G and Stoica, Ion and
             Gonzalez, Joseph E},
  journal = {arXiv preprint arXiv:2310.08560},
  year    = {2023}
}

@article{mei2024aios,
  title   = {{AIOS}: {LLM} Agent Operating System},
  author  = {Mei, Kai and Li, Zelong and Xu, Shuyuan and Ye, Ruosong
             and Ge, Yingqiang and Zhang, Yongfeng},
  journal = {arXiv preprint arXiv:2403.16971},
  year    = {2024}
}

@article{wu2024oscopilot,
  title   = {{OS-Copilot}: Towards Generalist Computer Agents with
             Self-Improvement},
  author  = {Wu, Zhiyong and Han, Chengcheng and Ding, Zichen and
             Weng, Zhenmin and Liu, Zhoumianze and Yao, Shunyu and
             Yu, Tao and Kong, Lingpeng},
  journal = {arXiv preprint arXiv:2402.07456},
  year    = {2024}
}

@article{liu2023lost,
  title   = {Lost in the Middle: How Language Models Use Long Contexts},
  author  = {Liu, Nelson F and Lin, Kevin and Hewitt, John and
             Paranjape, Ashwin and Bevilacqua, Michele and Petroni, Fabio
             and Liang, Percy},
  journal = {Transactions of the Association for Computational Linguistics},
  volume  = {12},
  pages   = {157--173},
  year    = {2024}
}

@article{schulman2017ppo,
  title   = {Proximal Policy Optimization Algorithms},
  author  = {Schulman, John and Wolski, Filip and Dhariwal, Prafulla
             and Radford, Alec and Klimov, Oleg},
  journal = {arXiv preprint arXiv:1707.06347},
  year    = {2017}
}

@article{belady1966,
  title   = {A Study of Replacement Algorithms for a Virtual-Storage Computer},
  author  = {Belady, L{\'a}szl{\'o} A},
  journal = {IBM Systems Journal},
  volume  = {5},
  number  = {2},
  pages   = {78--101},
  year    = {1966}
}

@article{sleator1985paging,
  title   = {Amortized Efficiency of List Update and Paging Rules},
  author  = {Sleator, Daniel D and Tarjan, Robert E},
  journal = {Communications of the ACM},
  volume  = {28},
  number  = {2},
  pages   = {202--208},
  year    = {1985}
}

@article{denning1968working,
  title   = {The Working Set Model for Program Behavior},
  author  = {Denning, Peter J},
  journal = {Communications of the ACM},
  volume  = {11},
  number  = {5},
  pages   = {323--333},
  year    = {1968}
}

@inproceedings{megiddo2003arc,
  title     = {{ARC}: A Self-Tuning, Low Overhead Replacement Cache},
  author    = {Megiddo, Nimrod and Modha, Dharmendra S},
  booktitle = {USENIX Conference on File and Storage Technologies (FAST)},
  pages     = {115--130},
  year      = {2003}
}

@inproceedings{kraska2018learned,
  title     = {The Case for Learned Index Structures},
  author    = {Kraska, Tim and Beutel, Alex and Chi, Ed H and Dean, Jeff
               and Polyzotis, Neoklis},
  booktitle = {Proceedings of the ACM SIGMOD International Conference on
               Management of Data},
  pages     = {489--504},
  year      = {2018}
}

@inproceedings{ge2024model,
  title     = {Model Tells You What to Discard: Adaptive {KV} Cache
               Compression for {LLMs}},
  author    = {Ge, Suyu and Zhang, Yunan and Liu, Liyuan and Zhang, Minjia
               and Han, Jiawei and Gao, Jianfeng},
  booktitle = {International Conference on Learning Representations},
  year      = {2024}
}

@inproceedings{liu2024cachegen,
  title     = {{CacheGen}: Fast Context Loading for Language Model Applications
               via {KV} Cache Streaming},
  author    = {Liu, Yuhan and Li, Hanchen and Du, Kuntai and Yao, Jiayi
               and Cheng, Yihua and Huang, Yuyang and Lu, Shan and
               Maire, Michael and Hoffmann, Henry and Holtzman, Ari and
               Jiang, Junchen},
  booktitle = {ACM SIGCOMM},
  year      = {2024}
}

@article{cybenko1989,
  title   = {Approximation by Superpositions of a Sigmoidal Function},
  author  = {Cybenko, George},
  journal = {Mathematics of Control, Signals and Systems},
  volume  = {2},
  number  = {4},
  pages   = {303--314},
  year    = {1989}
}

@article{hornik1991,
  title   = {Approximation Capabilities of Multilayer Feedforward Networks},
  author  = {Hornik, Kurt},
  journal = {Neural Networks},
  volume  = {4},
  number  = {2},
  pages   = {251--257},
  year    = {1991}
}

\end{document}